\documentclass[10pt,twocolumn,letterpaper]{article}

\usepackage[pagenumbers]{cvpr} 

\usepackage[dvipsnames]{xcolor}

\usepackage{booktabs}
\usepackage{multirow}
\usepackage{makecell}
\usepackage{tabularx}
\usepackage{xcolor}
\usepackage{colortbl}
\usepackage[flushleft]{threeparttable}
\usepackage{subcaption}
\usepackage{pifont}
\usepackage{colortbl}
\usepackage{bm}
\usepackage{graphicx}
\usepackage{setspace}
\usepackage{cuted}
\usepackage{soul}
\usepackage{dblfloatfix}
\usepackage[accsupp]{axessibility}
\newcommand{\cmark}{\ding{51}}

\definecolor{cvprblue}{rgb}{0.21,0.49,0.74}
\definecolor{cellgray}{gray}{0.9}
\definecolor{cellorange}{RGB}{251,229,214}
\definecolor{cellgreen}{RGB}{226,240,217}
\definecolor{cellyellow}{RGB}{255,242,204}
\definecolor{cellblue}{RGB}{222,235,247}

\usepackage[pagebackref,breaklinks,colorlinks,citecolor=cvprblue]{hyperref}

\newif\ifMain
\Maintrue  

\newif\ifAppendix
\Appendixtrue  

\def\confName{CVPR}
\def\confYear{2024}
\title{TAMM: TriAdapter Multi-Modal Learning for 3D Shape Understanding}

\author{%
Zhihao Zhang$^{1*}$\quad Shengcao Cao$^{2}$\thanks{\vspace{-5mm}Equal contribution.}\ \quad Yu-Xiong Wang$^{2}$\\
$^{1}$Xi'an Jiaotong University \quad
$^{2}$University of Illinois Urbana-Champaign\\
\texttt{$^{1}$zh1142@stu.xjtu.edu.cn \quad $^{2}$\{cao44,yxw\}@illinois.edu}
}
\begin{document}

\ifMain
\maketitle

\begin{strip}
  \centering
  \vspace{-8mm}
  \includegraphics[width= 1 \linewidth]{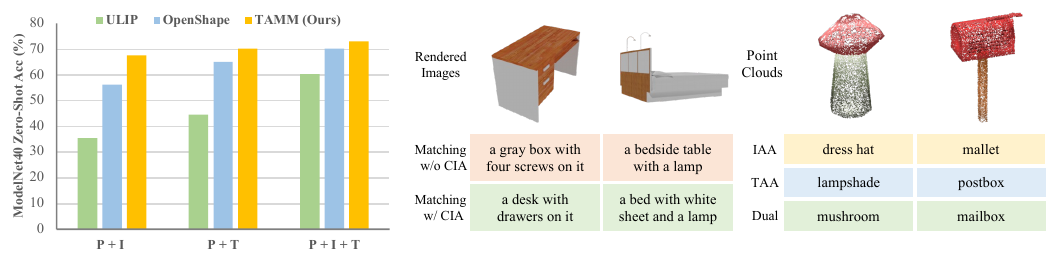}
  \captionof{figure}{\textbf{Our TriAdapter Multi-Modal Learning (TAMM) significantly enhances 3D shape understanding.} \textbf{Left:}
  When aligning features of 3D point clouds (P) with 2D images (I) and/or text (T),
  prior methods (\eg, ULIP~\cite{xue2023ulip} and OpenShape~\cite{liu2023openshape}) \emph{under-utilize the image modality}, due to the overlooked or unsolved image domain gap. TAMM better exploits the image modality and brings more gains when learning from both image and text data. The results are produced by pre-training Point-BERT~\cite{yu2022point} on ShapeNet~\cite{chang2015shapenet}. \textbf{Middle:} Our CLIP Image Adapter (CIA)
  \emph{re-aligns} the images rendered from 3D shapes with the text descriptions.
  The rendered images are
  {\sethlcolor{cellorange}\hl{inaccurately}}
  matched with text when the image features are directly extracted by CLIP, and CIA can {\sethlcolor{cellgreen}\hl{correct}} the matching.
  \textbf{Right:} Our Image Alignment Adapter (IAA) and Text Alignment Adapter (TAA) \emph{decouple 3D features} with complementary visual and semantic focuses. In the visualized examples, features from one single adapter are matched with classes whose {\sethlcolor{cellyellow}\hl{appearance}} or {\sethlcolor{cellblue}\hl{semantics}} resemble the true class; using both adapters leads to the  {\sethlcolor{cellgreen}\hl{correct}}
  class.
  }
  \label{fig:intro-teaser}
\end{strip}

\begin{abstract}
The limited scale of current 3D shape datasets hinders the advancements in 3D shape understanding, and motivates multi-modal learning approaches which transfer learned knowledge from data-abundant 2D image and language modalities to 3D shapes. However, even though the image and language representations have been aligned by cross-modal models like CLIP, we find that the image modality fails to contribute as much as the language in existing multi-modal 3D representation learning methods. This is attributed to the domain shift in the 2D images and the distinct focus of each modality. To more effectively leverage both modalities in the pre-training, we introduce TriAdapter Multi-Modal Learning (TAMM) -- a novel two-stage learning approach based on three synergistic adapters. First, our CLIP Image Adapter mitigates the domain gap between 3D-rendered images and natural images, by adapting the visual representations of CLIP for synthetic image-text pairs. Subsequently, our Dual Adapters decouple the 3D shape representation space into two complementary sub-spaces: one focusing on visual attributes and the other for semantic understanding, which ensure a more comprehensive and effective multi-modal pre-training. Extensive experiments demonstrate that TAMM consistently enhances 3D representations for a wide range of 3D encoder architectures, pre-training datasets, and downstream tasks. Notably, we boost the zero-shot classification accuracy on Objaverse-LVIS from 46.8\% to 50.7\%, and improve the 5-way 10-shot linear probing classification accuracy on ModelNet40 from 96.1\% to 99.0\%. Project page: \url{https://alanzhangcs.github.io/tamm-page}.
\end{abstract}
    
\vspace{-4mm}
\section{Introduction}
\label{sec:intro}

Despite the recent success of 3D shape understanding~\cite{yu2022point, ma2022rethinking, wu2022point, zhang2022pointclip,zhu2023pointclip, huang2023clip2point, hegde2023clip}, the limited scale of existing 3D shape datasets hinders the learning of more robust and generalizable 3D representations. 
Due to the significant human labor and expertise required in collecting and annotating 3D shape data, building large-scale 3D shape datasets is a prohibitive endeavor~\cite{chang2015shapenet, mo2019partnet, deitke2023objaverse}.

To mitigate the challenge of the limited data scale, recent research in 3D shape representation learning shows a promising direction of multi-modal learning~\cite{liu2023openshape, xue2023ulip}. By establishing connections among data of 3D shapes, 2D images, and text, it becomes possible to pre-train 3D shape representations by \emph{distilling learned knowledge from image and language modalities}. 
There exists a comprehensive body of literature of massive datasets~\cite{schuhmann2022laionb}, high-quality models pre-trained via self-supervised learning for 2D image~\cite{chen2020simple, he2020momentum, caron2021emerging} and language~\cite{devlin2019bert, brown2020language, touvron2023llama} modalities,
and furthermore, cross-modal vision-language models~\cite{radford2021learning, cherti2023reproducible}. For instance, ULIP~\cite{xue2023ulip} creates triplets of 3D point clouds, 2D images, and text by rendering images from 3D shapes and generating text descriptions from their metadata.
Then, ULIP adopts contrastive learning to align the 3D shape features with both the image features and text features extracted by CLIP~\cite{radford2021learning}. 

While the image and language modalities are pre-aligned by CLIP pre-training and share the same feature space, we observe that directly aligning 3D features with image features leads to \emph{considerably worse representation quality} as compared with aligning 3D features with text features, as shown in Figure~\ref{fig:intro-teaser}-left. This phenomenon suggests that existing methods (\eg, ULIP~\cite{xue2023ulip} and OpenShape~\cite{liu2023openshape}) are not optimally leveraging the 2D image modality in 3D representation pre-training.

There are two reasons for this \emph{counter-intuitive} observation: 1) The 2D images in the generated triplets follow a data distribution \emph{different from natural images}, on which CLIP is pre-trained. The images are rendered or projected from 3D to 2D, usually lacking a realistic background and texture. Since the image domain is shifted, the image features are no longer well-aligned with the text features, as exemplified in Figure~\ref{fig:intro-teaser}-middle. 2) Intuitively,
the \emph{image} features represent more \emph{visual} attributes including
the shape, texture, or color, while the \emph{text} features have a focus on \emph{semantics} such as the function of the object. As shown in Figure~\ref{fig:intro-teaser}-right, if the 3D features are specifically aligned with one single modality of images or text, they pay attention to different aspects of the 3D shape.
Therefore, enforcing the 3D shape to simultaneously align with two modalities that convey subtly distinct information could be challenging. Such an approach may not fully leverage the potential of multi-modal learning signals.

Aiming at addressing these two issues hindering multi-modal pre-training for 3D shape understanding, we propose \textbf{T}ri\textbf{A}dapter \textbf{M}ulti-\textbf{M}odal Learning (\textbf{TAMM}), a two-stage pre-training approach based on three synergistic adapters.
In the first stage, to mitigate the domain gap between the 2D images rendered from 3D shapes and the natural images on which CLIP is pre-trained, we adapt the visual representations of CLIP based on the synthetic image-text pairs. More specifically, we fine-tune a lightweight CLIP Image Adapter on top of CLIP visual encoder through contrastive learning and re-align the adapted image features with the text features in the new domain. This CLIP Image Adapter allows us to establish more accurate relations between 3D shapes, 2D images, and language in an updated feature space, and avoids learning 3D representations from mismatched image features and text features.

In the following second stage, to prevent the vision-semantic feature disparity from impairing our 3D representation pre-training, we choose to embrace this disparity and decouple the 3D representations into two sub-spaces.
To comprehensively encode a 3D shape, the 3D encoder needs to capture both the visual and semantic aspects of its representation, which are centric to the corresponding image and text features, respectively. Therefore, we decouple the 3D feature space for the two focuses on visual and semantic representations.
In particular, we attach two independent feature adapters to the 3D backbone and transform the 3D features into two sub-spaces. One sub-space focuses more on the visual representations, and is aligned with the 2D image feature space; the other sub-space focuses more on the semantic representations, and is aligned with the language feature space. This approach of decoupled feature spaces
makes the learned 3D representations more comprehensive and expressive.

To summarize, our main contribution includes:
\begin{itemize}[leftmargin=*, noitemsep, nolistsep]
\item We identify the under-utilization of the 2D image modality in existing multi-modal methods. The image domain gap and feature disparity in image-text pairs hinder representation learning in 3D shape understanding.
\item We propose a novel multi-modal learning framework with two learning stages and three unified adapter modules. Our proposed TAMM better exploits both image and language modalities and improves 3D shape representations.
\item Our TAMM consistently enhances 3D representations for a variety of 3D encoder architectures (\eg, Point-BERT~\cite{yu2022point}, SparseConv~\cite{choy20194d}), pre-training datasets (\eg, ShapeNet~\cite{chang2015shapenet}, an ensembled dataset~\cite{liu2023openshape}), and downstream tasks (\eg, zero-shot and linear probing shape classification on Objaverse-LVIS~\cite{deitke2023objaverse}, ModelNet40~\cite{wu20153d}, and ScanObjectNN~\cite{uy2019revisiting}).
\end{itemize}

\section{Related Work}
\label{sec:relatted}
\noindent\textbf{3D Shape Understanding.}
There are two mainstreams for 3D shape representation learning: 1) Projecting 3D shapes into voxel or grid-based formats~\cite{shi2020pv} and then using 2D/3D convolutions~\cite{choy20194d} for feature extraction. 2) Directly modeling 3D point clouds with point-centric architectures~\cite{qi2017pointnet, qi2017pointnet++, zhang2021self, qian2022pointnext, wu2022point, ma2022rethinking, pang2022masked, zhang2022point}.
In this work, to ensure a fair and comprehensive comparison with previous methods~\cite{xue2023ulip, liu2023openshape} on the pre-training scheme,
we follow their selection and utilize two representative 3D encoders from these two mainstreams: SparseConv~\cite{choy20194d} and Point-BERT~\cite{yu2022point}. SparseConv is designed for efficiently processing sparse voxels using specialized convolutions that focus computations on non-zero data points. Point-BERT~\cite{wu2022point} utilizes a Transformer-based architecture~\cite{vaswani2017attention} and can be self-supervised by masked modeling~\cite{devlin2019bert}.

\noindent\textbf{Multi-Modal Representation Learning.}
Contrastive Language-Image Pre-training (CLIP)~\cite{radford2021learning} has enabled various downstream applications including object detection~\cite{zareian2021open, zhong2022regionclip, zhou2022detecting} and language grounding~\cite{li2022grounded}.   
Recently, CLIP has been extended to 3D-based tasks, such as zero-shot text-to-3D generation~\cite{hong2022avatarclip, tevet2022motionclip, mohammad2022clip, jain2022zero, michel2022text2mesh, xu2023dream3d} and scene-level 3D segmentation~\cite{ha2022semantic, peng2023openscene, yang2023regionplc, rozenberszki2022language}. Meanwhile, developing general and robust representations for 3D shape understanding with the foundation of CLIP~\cite{zhang2022pointclip, huang2023clip2point, zhu2023pointclip, xue2023ulip, liu2023openshape, zhou2023uni3d} becomes a major focus. Among these methods, ULIP~\cite{xue2023ulip}, as a pioneering work, utilizes contrastive learning to distill CLIP features into 3D representations. OpenShape~\cite{liu2023openshape} follows this learning paradigm with a focus on building a larger pre-training dataset with enriched and filtered text data. Unlike OpenShape~\cite{liu2023openshape}, we focus on improving the multi-modal learning paradigm by more effectively leveraging both the image and text modalities via a two-stage pre-training approach based on three synergistic adapters.

\begin{figure*}
  \centering
  \vspace{-4mm}
  \includegraphics[width= 1 \linewidth]{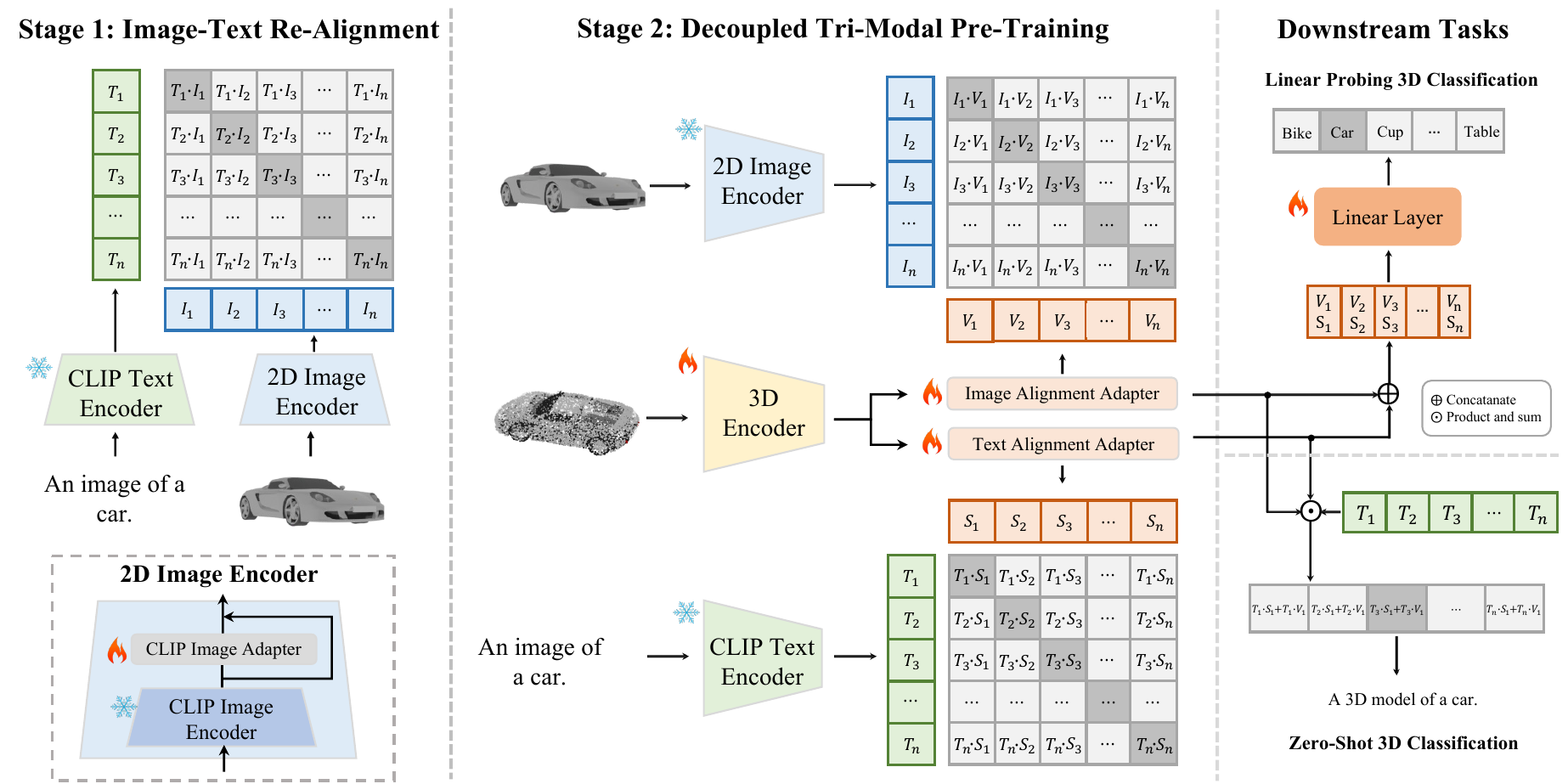}
  \caption{\textbf{Overview of TAMM.} \textbf{Left:} In Stage~1, TAMM fine-tunes a lightweight CLIP Image Adapter (CIA) through contrastive learning and re-aligns the image features with the text features to alleviate the domain shift originated from rendered images. Contrastive learning maximizes inner products between features from corresponding text-image pairs, and reduces similarities of mismatched pairs. \textbf{Middle:} In Stage~2, TAMM introduces Image Alignment Adapter (IAA) and Text Alignment Adapter (TAA) to decouple 3D representations into two sub-spaces: one focusing more on visual attributes and the other for semantic understanding, ensuring a more comprehensive and effective multi-modal pre-training strategy.
  \textbf{Right:} TAMM adaptively utilizes decoupled 3D features for various downstream tasks including linear probing classification (top) and zero-shot classification (bottom), achieving more robust classification results. 
  }
  \label{sec:method-model-arch}
  \vspace{-2mm}
\end{figure*}

\section{Method}
\label{sec:method}
Figure~\ref{sec:method-model-arch} shows our proposed two-stage approach, \textbf{T}ri\textbf{A}dapter \textbf{M}ulti-\textbf{M}odal Learning (\textbf{TAMM}), for pre-training robust and generalizable 3D representations by leveraging 2D images and text. 
We first revisit the triplet data generation and problem formulation in multi-modal 3D representation learning.
Section~\ref{sec:method-clip-tuning} delves into our fine-tuning of the original CLIP model for better fitting 3D understanding tasks. Section~\ref{sec:method-multi-modal} details our strategy to enhance alignment across the 3D, 2D, and text feature spaces.

\noindent\textbf{Problem Formulation.}
\label{sec:method-openshape}
Given $n$ triplets $\{(P_i, I_i, T_i)\}_{i=1}^n$, where $P_i$ is a 3D point cloud, $I_i$ represents the corresponding image produced by projecting the 3D point cloud $P_i$ into 2D from an arbitrary perspective, and $T_i$ denotes the associated text generated using advanced vision-language models such as BLIP~\cite{li2022blip}, the objective is to learn high-quality 3D representations from the triplets. 
The basic framework to achieve this objective is proposed by ULIP~\cite{xue2023ulip} (and followed by OpenShape~\cite{liu2023openshape}) with the help of CLIP~\cite{radford2021learning, cherti2023reproducible}.

Formally, the 3D feature $f_i^P = E_P(P_i)$ is produced by a learnable 3D encoder $E_P$, and the corresponding image feature $f_i^I = E_I(I_i)$ and text feature $f_i^T = E_T(T_i)$ are generated by frozen CLIP encoders $E_I,E_T$. Then the 3D encoder $E_P$ is optimized by aligning the 3D feature space $f^P$ simultaneously to the pre-aligned CLIP image space $f^I$ and text space $f^T$ through contrastive learning. The corresponding contrastive loss $L_\text{contrast}(f^{M_1}, f^{M_2})$ between two modalities $M_1,M_2$ (3D-2D or 3D-text) is formulated as:
\begin{equation}
\label{equation-contrastive_loss}
\scriptsize
 - \frac{1}{2n} \sum\limits_{i=1}^{n} \left( \log\frac{\exp(f_i^{M_1} \cdot f_i^{M_2}/ \tau)}{\sum\limits_{j=1}^{n} \exp(f_i^{M_1} \cdot f_j^{M_2} / \tau)} +  \log\frac{\exp(f_i^{M_2} \cdot f_i^{M_1}/ \tau)}{\sum\limits_{j=1}^{n}\exp(f_i^{M_2} \cdot f_j^{M_1} / \tau)} \right),
\end{equation}
where $\tau$ denotes the temperature hyperparameter.

\subsection{Image-Text Re-Alignment}
\label{sec:method-clip-tuning}
Unlike ULIP~\cite{xue2023ulip} or OpenShape~\cite{liu2023openshape}, in which the 3D feature space is aligned with the image-text feature spaces \emph{directly produced by CLIP}, we introduce an image feature space tuning strategy, aiming to \emph{foster better alignment} across 3D, 2D, and language modalities (Figure~\ref{sec:method-model-arch}-left).
We argue that the image feature space produced directly by the CLIP visual encoder does not perfectly align with the text feature space and is thus sub-optimal. The reason is that the 2D images in the triplets, which originate from 3D point cloud projections, lack backgrounds and are visually different from natural images on which CLIP is pre-trained.

Therefore, when using such 2D images from a shifted data domain to perform multi-modal pre-training, it becomes necessary to further fine-tune CLIP and re-align its image and language feature spaces.
For the first time, we re-design domain adapters~\cite{houlsby2019parameter} for multi-modal contrastive 3D representation learning, and propose a \textbf{C}LIP \textbf{I}mage \textbf{A}dapter (\textbf{CIA}), to adapt the image feature space for the rendered 2D images in our data triplets. We append a lightweight, learnable multi-layer perceptron (MLP) to CLIP image encoder, with a residual connection which seamlessly integrates the new knowledge acquired from fine-tuning with the existing knowledge from the pre-trained CLIP backbone. To avoid heavy computation and over-fitting, we only fine-tune the additional parameters in CIA, instead of the whole CLIP backbone. Formally, given the image feature $f_i^I$ and text feature $f_i^T$ extracted from the triplet $(P_i, T_i, I_i)$,
we use a learnable, two-layer CIA $A_C(\cdot)$ to adapt the image feature, formulated as: 
\begin{equation}
A_C(f_i^I) = \sigma(f_i^IW_1) \cdot W_2,
\end{equation}
where $W_1$ and $W_2$ are the parameters associated with the linear transformation layers, and $\sigma$ is the non-linear activation function. The refined image feature $\widetilde f_i^I$ can be computed with a residual connection: 
\begin{equation}
\label{equation-1}
\widetilde f_i^I = \alpha A_C(f_i^I) + (1 - \alpha) f_i^I,
\end{equation}
where $\alpha$ is a hyperparameter.
Finally, CIA $A_C(\cdot)$
is optimized by minimizing the contrastive loss function (Equation~\ref{equation-contrastive_loss}), instantiated as:
\begin{equation}
\mathcal{L}_\text{realign} = L_\text{contrast}(\widetilde f^I, f^T ).
\end{equation}

\subsection{Decoupled Tri-Modal Pre-Training}
\label{sec:method-multi-modal}
Although the image and language feature spaces have been aligned by CLIP and our CLIP Image Adapter, they still encode \emph{subtly different information}. For instance, the color of a 3D shape may be inferred from the image feature, but it cannot possibly be included in the text feature if the text description does not contain color information. Similarly, the image feature may not include semantic information such as the function or name of an object. In such cases, enforcing the 3D feature to align with both the image and text features \emph{simultaneously} is challenging.

In order to overcome the obstacle introduced by aligning the 3D shape feature space with two distinct modalities, we propose a novel decoupled tri-modal pre-training framework, which aligns two decoupled 3D shape feature spaces with the refined image feature space and the text feature space, respectively (Figure~\ref{sec:method-model-arch}-middle). We encourage the 3D encoder to cover both the \emph{visual} representation and the \emph{semantic} information inherent to the 3D shape, avoiding the dilemma of aligning with disparate image and text features at the same time.
Formally, given the triplets $\{(P_i, I_i, T_i)\}_{i=1}^n$, we first generate the image feature $\widetilde f_i^I$ using the frozen, adapted 2D image encoder (Equation \ref{equation-1}) and the text feature $f_i^T$ using the original CLIP text encoder. 
We introduce a pair of lightweight Dual Adapters: \textbf{I}mage \textbf{A}lignment \textbf{A}dapter (\textbf{IAA}) $A_V(\cdot)$ and \textbf{T}ext \textbf{A}lignment \textbf{A}dapter (\textbf{TAA}) $A_S(\cdot)$. They split the 3D feature $f_i^P$ originated from the 3D encoder $E_P$ into a \textbf{v}ision-focusing feature $f_i^{VP}$ and a \textbf{s}emantic-focusing feature $f_i^{SP}$, respectively.
Dual Adapters $A_V(\cdot)$ and $A_S(\cdot)$ share the same architecture as the CLIP Image Adapter in Section~\ref{sec:method-clip-tuning} and can be formulated as:
\begin{equation}
\begin{split}
f_i^{VP} = A_V(f_i^P) = \sigma(f_i^PW_1^V) \cdot W_2^V,\\
f_i^{SP} = A_S(f_i^P) = \sigma(f_i^PW_1^S) \cdot W_2^S,
\end{split}
\end{equation}
where $W_1^V$, $W_2^V$, $W_1^S$, and $W_2^S$ are the parameters associated with the linear layers, and $\sigma$ represents the activation function. 
By decoupling 3D features into these sub-spaces, the 3D encoder $E_P$ interprets the 3D shape with a more comprehensive visual and semantic understanding, improving the expressivity of the learned representations. 

Finally, instead of enforcing the 3D shape feature space to directly mimic the pre-aligned image-text feature space, we use the decoupled 3D features $f^{VP}, f^{SP}$ and align them with the adapted image feature space $\widetilde f^I$ and the text feature space $f^T$, respectively.  
Moreover, since one single image can only capture the 3D shape from one perspective, we align the 3D feature with the adapted features of \emph{multi-view images}, to fully exploit the image modality and achieve a better alignment between 3D and 2D.
The overall loss function is defined as:
\begin{equation}
\footnotesize
\mathcal{L}_\text{trimodal} = L_\text{contrast}(f^{SP}, f^T) + \frac{1}{m}\sum_{k}^{m} L_\text{contrast}(f^{VP}, \widetilde{f}^{I, k}),
\end{equation}
where $m$ represents the number of rendered images and $\widetilde{f}^{I, k}$ is the adapted image feature from the $k$-th view.

\noindent\textbf{Application in Downstream Tasks.} The learned 3D feature sub-spaces, $f^{VP}$ and $f^{SP}$ can be adaptively applied to a variety of downstream tasks (Figure~\ref{sec:method-model-arch}-right).
Specifically, in zero-shot 3D classification, we leverage both the 3D vision-focusing feature $f^{VP}$ and the 3D semantic-focusing feature $f^{SP}$. We calculate the similarity between these features and category embeddings generated by the CLIP text encoder, respectively. After summing up the per-category similarity scores, the category with the highest similarity is chosen as the predicted class, yielding more robust and enhanced classification results compared with using a single sub-space alone. 
In the linear probing classification task, we concatenate the 3D vision-focusing feature $f^{VP}$ and 3D semantic-focusing feature $f^{SP}$ as input to the learnable linear classification layer, providing a more comprehensive and robust representation for 3D understanding.

\section{Experiments}
\label{sec:experiment}

\begin{table*}[!t]
\centering
\vspace{-4mm}
\resizebox{0.95 \textwidth}{!}{
\begin{tabular}{cc|c|ccc|ccc|ccc}
\toprule
\multirow{2}{*}{\makecell[c]{Pre-Training \\ Dataset}} &\multirow{2}{*}{Model}&\multirow{2}{*}{\makecell[c]{Pre-Training \\ Method}}&\multicolumn{3}{c|}{Objaverse-LVIS~\cite{deitke2023objaverse}} &\multicolumn{3}{c|}{ModelNet40~\cite{wu20153d}} & \multicolumn{3}{c}{ScanObjectNN~\cite{uy2019revisiting}} \\
&&&Top-1&Top-3&Top-5&Top-1&Top-3&Top-5&Top-1&Top-3&Top-5 \\
\midrule
- &PointCLIP~\cite{zhang2022pointclip} & - &1.9&4.1&5.8&19.3&28.6&34.8& 10.5 & 20.8& 30.6 \\ 
- & PointCLIP v2~\cite{zhu2023pointclip}&- &4.7&9.5&12.9&63.6&77.9&85.0& 42.2& 63.3& 74.5 \\
\midrule
\multirow{7}{*}{ShapeNet}  & ViT-B/32~\cite{dosovitskiy2020image} & CLIP2Point~\cite{huang2023clip2point} & 2.7 & 5.8 & 7.9 & 49.5& 71.3 & 81.2 &25.5 & 44.6 & 59.4\\
& Transformer~\cite{vaswani2017attention} & ReCon~\cite{qi2023contrast} & 1.1 & 2.7& 3.7& 61.2& 73.9& 78.1& 42.3& 62.5& 75.6\\
\cmidrule(lr){2-12}

& \multirow{2}{*}{SparseConv~\cite{choy20194d}} & OpenShape~\cite{liu2023openshape} & 11.6 & 21.8 & 27.1 & 72.9 & 87.2 & 93.0 & 52.7 & 72.7 & 83.6\\
& & TAMM (Ours) & 13.6 & 24.2 & 29.3 & 74.6 & 88.2 & 94.0 & 57.9 & 75.3 & 83.1\\
\cmidrule(lr){2-12}

& \multirow{3}{*}{Point-BERT~\cite{yu2022point}} & ULIP~\cite{xue2023ulip} & 6.2 & 13.6 & 17.9 & 60.4 & 79.0 & 84.4 &51.5 &71.1 & 80.2\\
& & OpenShape~\cite{liu2023openshape} & 10.8 & 20.2 & 25.0 & 70.3 & 86.9 & 91.3 & 51.3 & 69.4 & 78.4\\
& & TAMM (Ours) & 13.7 & 24.2 & 29.2 & 73.1 & 88.5 & 91.9 & 54.8 & 74.5 & 83.3\\

\midrule
\multirow{5}{*}{\makecell[c]{Ensembled \\ (no LVIS)}}  & \multirow{2}{*}{SparseConv~\cite{choy20194d}} &OpenShape~\cite{liu2023openshape} & 37.0 & 58.4 & 66.9 & 82.6 & 95.0 & 97.5 & 54.9 & 76.8 & 87.0 \\
& & TAMM (Ours) &39.8 &62.0&70.4&85.7&\textbf{96.8}&\textbf{98.3} & 57.5 & 81.3 & 90.0
\\ 
\cmidrule(lr){2-12}
& \multirow{3}{*}{Point-BERT~\cite{yu2022point}} & ULIP~\cite{xue2023ulip} & 21.4 & 38.1 & 46.0 & 71.4 & 84.4 & 89.2 & 46.0 & 66.1 & 76.4\\ 
& & OpenShape~\cite{liu2023openshape} & 39.1 & 60.8 & 68.9 & 85.3 & 96.2 & 97.4 & 47.2 & 72.4 & 84.7  \\
& & TAMM (Ours) &42.0 & 63.6& 71.7& \textbf{86.3}& 96.6&98.1 & 56.7 & 78.3 & 86.1 \\

\midrule

 \multirow{5}{*}{Ensembled} &\multirow{2}{*}{SparseConv~\cite{choy20194d}} &OpenShape~\cite{liu2023openshape} & 43.4 & 64.8 & 72.4 & 83.4 & 95.6 & 97.8 & 56.7 & 78.9 & 88.6 \\
& & TAMM (Ours) & 43.8 & 66.2 & 74.1& 85.4& 96.4& 98.1& \textbf{58.5}& \textbf{81.3}&\textbf{89.5} \\ 
\cmidrule(lr){2-12}
 & \multirow{3}{*}{Point-BERT~\cite{yu2022point}}  & ULIP~\cite{xue2023ulip} & 26.8 & 44.8 & 52.6 & 75.1 & 88.1 & 93.2 & 51.6 & 72.5 & 82.3\\ 
& & OpenShape~\cite{liu2023openshape} & 46.8 & 69.1 & 77.0 & 84.4 & 96.5 & 98.0 & 52.2 & 79.7 & 88.7 \\
& & TAMM (Ours) & \textbf{50.7} & \textbf{73.2} & \textbf{80.6} & 85.0 & 96.6 & 98.1 & 55.7 & 80.7 & 88.9 \\
\bottomrule
\end{tabular}}
\caption{\textbf{Zero-shot 3D classification results.}  TAMM sets new state of the art in zero-shot classification accuracy across Objaverse-LVIS, ModelNet-40, and ScanObjectNN benchmarks, outperforming existing methods in diverse settings of pre-training datasets and 3D model architectures. The performance gain is more significant on the most challenging long-tailed Objaverse-LVIS dataset.
}
\label{sec:experiment-zsd_1}
\vspace{-2mm}
\end{table*}
\label{sec:experiment-datasets}
\noindent \textbf{Pre-Training Datasets.}
Following the prior state-of-the-art method, OpenShape~\cite{liu2023openshape},   our TAMM is pre-trained on the triplets generated from four datasets: ShapeNetCore~\cite{chang2015shapenet}, 3D-FUTURE~\cite{fu20213d}, ABO~\cite{collins2022abo}, and Objaverse~\cite{deitke2023objaverse}. 
Our training sets are defined as follows: 
``ShapeNet'' is a triplet set derived exclusively from the ShapeNetCore dataset, containing $52,470$ 3D shapes with corresponding images and text;
``Ensembled (no LVIS)'' is a set of $829,460$ triplets from the above datasets excluding Objaverse-LVIS; 
``Ensembled'' denotes the triplet set comprising data from all four datasets, containing $875,665$ 3D shapes and their associated images and text.

\noindent \textbf{Evaluation Datasets.}
Our TAMM is evaluated on the following datasets:  Objaverse-LVIS~\cite{deitke2023objaverse}, ModelNet40~\cite{wu20153d}, ScanObjectNN~\cite{uy2019revisiting}, and ScanNet~\cite{dai2017scannet}. Objaverse-LVIS encompasses a broad range of categories, featuring $46,832$ high-quality shapes distributed across $1,156$ LVIS~\cite{gupta2019lvis} categories. ModelNet40 is a synthetic indoor 3D dataset comprising $40$ categories. We use the test split of $2,468$ shapes in our experiments. ScanObjectNN consists of scanned objects from $15$ common categories. It has three main variants: OBJ-BG, OBJ-ONLY, and PB-T50-RS. ScanNet, characterized by its real-world scans, includes 1,513 indoor scenes containing 36,213 objects. We conduct experiments on four tasks including zero-shot 3D classification, linear probing 3D classification, few-shot linear probing 3D classification, and real-world recognition, to demonstrate the advantages of our TAMM. Other implementation details regarding pre-training and evaluation are introduced in the supplementary material.

\subsection{Zero-Shot 3D Classification}
\label{sec:experiment:zero-shot-subsection}
Zero-shot ability is a key metric for reflecting the quality of learned 3D representations, which requires the representations to be \emph{directly applicable to datasets where the model has never been explicitly supervised}. Without any further tuning, the 3D representations are compared with text embeddings of categories to predict the classes of 3D shapes.
To make a fair comparison and keep consistency with prior work, we adopt the same settings as OpenShape~\cite{liu2023openshape} on three benchmarks: Objaverse-LVIS~\cite{deitke2023objaverse}, ModelNet40~\cite{wu20153d}, and OBJ-ONLY (ScanObjectNN)~\cite{uy2019revisiting}.

The results are summarized in Table~\ref{sec:experiment-zsd_1}. First, we observe that TAMM, benefited from multi-modal pre-training, outperforms PointCLIP~\cite{zhang2022pointclip} and PointCLIP v2~\cite{zhu2023pointclip} by a large margin. 
Furthermore, it is evident that our pre-trained models consistently outperform those pre-trained by ULIP and OpenShape, \emph{irrespective of} whether they are pre-trained on smaller datasets like ShapeNet or more expansive ones such as the Ensembled dataset. 
For instance, Point-BERT~\cite{yu2022point} pre-trained by our TAMM on the Ensembled (no LVIS) dataset surpasses ULIP and OpenShape by margins of $+20.6\%$ and $+2.9\%$ in Top-1 accuracy on the long-tailed Objaverse-LVIS benchmark, which validates the effectiveness of our multi-modal pre-training scheme.

{
\setlength\tabcolsep{3pt}
\begin{table}[t]
\centering

\resizebox{0.48\textwidth}{!}{%
\begin{tabular}{cc|c|c|ccc}
\toprule
Pre-Training & \multirow{2}{*}{Method}& O-LVIS & M-40& \multicolumn{3}{c}{ScanObjectNN~\cite{uy2019revisiting}} \\
Dataset && \cite{deitke2023objaverse} & \cite{wu20153d} & \scriptsize OBJ-BG & \scriptsize OBJ-ONLY & \scriptsize PB-T50-RS \\
\midrule

\multirow{4}{*}{ShapeNet} & ULIP~\cite{xue2023ulip} & 34.6 & 90.6& 75.4 & 75.4 & 64.8 \\
& OpenShape~\cite{liu2023openshape} &29.3 & 88.5 & 77.8 & 78.5 & 64.1 \\
& TAMM (Ours) & 39.1& 91.0& 80.6 & 81.1 & 68.5 \\
& Rel. Improv. & +9.8 & +2.5 & +2.8 & +2.6 & +4.4 \\
\midrule
\multirow{3}{*}{Ensembled}& OpenShape~\cite{liu2023openshape} & 48.3 & 91.3 & 85.9 & 85.4 & 78.0 \\

& TAMM (Ours) & \textbf{59.5}& \textbf{93.5}& \textbf{88.5} &\textbf{88.0} & \textbf{80.3} \\
& Rel. Improv. & +11.2 & +2.2 & +2.6 & +2.6 & +1.7 \\
\bottomrule

\end{tabular}}

\caption{\textbf{Linear probing 3D classification results.} TAMM outperforms previous methods by a large margin, \eg, $+11.2\%$ accuracy gain on the challenging Objaverse-LVIS dataset.}
\label{sec:experiment-linear_probing_scanobjectnn}
\vspace{-2mm}
\end{table}
}

\subsection{Linear Probing 3D Classification}
\label{sec:experiment:linear-probing-subsection}
To further evaluate the 3D representation quality of our pre-trained models, we design and conduct linear probing classification experiments. Here, we first freeze our pre-trained Point-BERT model along with the Image Alignment Adapter (IAA) and Text Alignment Adapter (TAA), and then append a single learnable linear layer to the model. Given a batch of point clouds, as illustrated in Figure~\ref{sec:method-model-arch}, we generate features from both IAA and TAA and concatenate them, and learn the appended linear classification layer on the concatenated features.
We evaluate the accuracy of the linear classifier on three benchmarks: Objaverse-LVIS~\cite{deitke2023objaverse}, ModelNet40~\cite{wu20153d}, and ScanObjectNN~\cite{uy2019revisiting}.
Objaverse-LVIS is a challenging long-tailed dataset with $1,156$ categories, which has not been evaluated by prior methods~\cite{xue2023ulip, liu2023openshape}. To ensure evaluation with statistically meaningful number of samples per class, we exclude categories with fewer than $10$ instances, leaving us $1,046$ categories. Samples in each class are divided into training and testing sets at an $8:2$ ratio. For ModelNet40 and ScanObjectNN, we use the standard splits, following~\cite{wu2022point, qi2023contrast}.

The results are shown in Table~\ref{sec:experiment-linear_probing_scanobjectnn}.
TAMM consistently outperforms all previous methods by a large margin. For example, pre-trained on the Ensembled dataset, TAMM improves over OpenShape by $+11.2\%$ and $+2.2\%$ overall accuracy on Objaverse-LVIS and ModelNet40, respectively. 
This demonstrates that TAMM effectively exploits knowledge from CLIP and learns generalizable 3D representations that are \emph{directly applicable to novel linear classification tasks}.
Notably, TAMM even surpasses the Point-BERT model pre-trained without multi-modality but allowed to be \emph{fully fine-tuned} on OBJ-BG ($87.4\%$)~\cite{yu2022point} by $+1.1\%$ accuracy.
These strong results indicate that TAMM has learned excellent transferable 3D representations, showing a great potential in real-world applications. 

\subsection{Few-Shot Linear Probing 3D Classification}
Following the previous evaluation, we perform few-shot classification experiments on ModelNet40~\cite{wu20153d} to assess TAMM in low-data scenarios. Similar to the linear probing experiments described in Section~\ref{sec:experiment:linear-probing-subsection}, we extend our model with an additional linear classification layer and only train this linear layer instead of fine-tuning the entire model. 
Following previous work~\cite{yu2022point, zhang2022point}, 
we adopt the ``$K$-way $N$-shot'' configuration, wherein we select $K$ classes at random and sample ($N+20$) instances per class. 
Training is conducted on a \emph{support set} of $K \times N$ samples, while evaluation is based on a \emph{query set} comprised of the remaining $20$ instances per class. 
We assess our model under four distinct scenarios: ``$5$-way $10$-shot,'' ``$5$-way $20$-shot,'' ``$10$-way $10$-shot,'' and ``$10$-way $20$-shot.'' For each scenario, we carry out $10$ separate trials, and report both the mean performance and the standard deviation across these trials.

{
\setlength\tabcolsep{3pt}
\begin{table}[t]
\centering
\resizebox{0.48\textwidth}{!}{%
\begin{tabular}{cc|cccc}
\toprule
Pre-Training & \multirow{2}{*}{Method}& \multicolumn{2}{c}{5-way} & \multicolumn{2}{c}{10-way} \\
Dataset && \small 10-shot & \small 20-shot & \small 10-shot & \small 20-shot \\
\midrule
\multirow{3}{*}{ShapeNet} & ULIP~\cite{xue2023ulip} & $94.4_{\pm 3.7}$ & $93.2_{\pm 4.2}$ & $86.6_{\pm 5.3}$ & $90.6_{\pm 5.2}$ \\
& OpenShape~\cite{liu2023openshape} & $95.3_{\pm 2.6}$ & $97.9_{\pm 3.9}$ & $89.2_{\pm 5.1}$ & $92.9_{\pm 3.9}$ \\
& TAMM (Ours) & $97.8_{\pm 1.9}$ & $98.1_{\pm 1.4}$ & $95.3_{\pm 3.9}$ & $95.9_{\pm 2.6}$ \\
\midrule
\multirow{2}{*}{Ensembled}&OpenShape~\cite{liu2023openshape} & $96.1_{\pm 2.7}$ & $95.7_{\pm 2.5}$ & $89.1_{\pm 4.6}$ & $91.8_{\pm 3.7}$ \\

& TAMM (Ours) & $\bm{99.0_{\pm 1.3}}$ & $\bm{99.4_{\pm 0.7}}$ & $\bm{96.8_{\pm 2.9}}$ & $\bm{97.4_{\pm 2.2}}$ \\
\bottomrule
\end{tabular}
}

\caption{\textbf{Few-shot linear probing classification results on ModelNet40.} We report the average accuracy and standard deviation of 10 independent experiments. Our TAMM consistently achieves both the best average accuracy and the lowest variance in various few-shot settings.}
\label{sec:experiment-few-shot-linear}
\vspace{-2mm}
\end{table}
}

{
\setlength\tabcolsep{3pt}
\begin{table*}[ht]
\centering
\vspace{-4mm}
\resizebox{1 \textwidth}{!}{
\begin{tabular}{c|c|ccccccccccccccccc}
\toprule
Method & Avg. & Bed & Cab & Chair & Sofa & Tabl & Door & Wind & Bksf & Pic & Cntr & Desk & Curt & Fridg & Bath & Showr & Toil & Sink \\
\midrule
CLIP2Point~\cite{huang2023clip2point} & 24.9 & 20.8 & 0.0 & \textbf{85.1} & 43.3 & 26.5 & \textbf{69.9} & 0.0 & 20.9 & 1.7 & 31.7 & 27.0 & 0.0 & 1.6 & 46.5 & 0.0 & 22.4 & 25.6 \\
PointCLIP w/ TP.~\cite{zhu2023pointclip} & 26.1 & 0.0 & 55.7 & 72.8 & 5.0 & 5.1 & 1.7 & 0.0 & 77.2 & 0.0 & 0.0 & 51.7 & 0.3 & 0.0 & 0.0 & 40.3 & 85.3 & \textbf{49.2} \\

CLIP2Point w/ TP.~\cite{huang2023clip2point} & 35.2 & 11.8 & 3.0 & 45.1 & 27.6 & 10.5 & 61.5 & 2.6 & 71.9 & 0.3 & \textbf{33.6} & 29.9 & 4.7 & \textbf{11.5} & \textbf{72.2} & \textbf{92.4} & \textbf{86.1} & 34.0 \\
CLIP$^2$~\cite{zeng2023clip2} & 38.5 & 32.6 & \textbf{67.2} & 69.3 & 42.3 & 18.3 & 19.1 & 4.0 & 62.6 & 1.4 & 12.7 & 52.8 & 40.1 & 9.1 & 59.7 & 41.0 & 71.0 & 45.5 \\

OpenShape\textsuperscript{$\dagger$}~\cite{liu2023openshape} & 45.6 & \textbf{66.7} & 3.2 & 75.8 & 83.5 & 37.7 & 49.2 & 47.5 & 64.9 &  48.2 & 1.9 &  \textbf{66.1} & 70.2 & 1.8  & 50.0 & 57.1 & 45.2 & 7.1 \\
\midrule
TAMM (Ours)\textsuperscript{$\dagger$} & \textbf{49.4} & \textbf{66.7} & 4.8 & 83.6 & \textbf{84.5} & \textbf{48.9} & 57.9 & \textbf{48.2} & \textbf{80.5} & \textbf{61.3} & 1.9 & 60.6 & \textbf{83.6} & 7.0  & 41.4 & 56.1 & 48.4 & 3.6\\
\bottomrule
\end{tabular}
}

\begin{spacing}{0.8}
{\footnotesize w/ TP. denotes training with the real-world data provided by CLIP$^2$.}
{\footnotesize \textsuperscript{$\dagger$} Results using Point-BERT~\cite{yu2022point} as 3D encoder, pre-trained on the Ensembled dataset.}
\end{spacing}
\caption{\textbf{Zero-shot classification results on the real-world ScanNet dataset. Avg.}: Mean average Top-1 accuracy of all classes. TAMM achieves the best results.}
\label{sec:experiment-open-world-understanding}
\vspace{-2mm}
\end{table*}
}

{
\setlength\tabcolsep{2pt}
\begin{table*}[t]
\centering
\begin{subtable}[t]{0.4\textwidth}
\centering
\resizebox{\textwidth}{!}{%
\begin{tabular}{@{}ccc|c|ccc|ccc@{}}
\multirow{2}{*}{CIA} & \multirow{2}{*}{IAA} & \multirow{2}{*}{TAA} & \multirow{2}{*}{\makecell[c]{Img-Txt \\ Acc}} &\multicolumn{3}{c|}{Objaverse-LVIS} & \multicolumn{3}{c}{ModelNet-40} \\
&&&& Top-1 &Top-3& Top-5 & Top-1 &Top-3& Top-5\\
\midrule
& \cmark & \cmark &40.1& 13.5 & 23.9& 29.1& 72.8& 88.2& 91.7\\
\cmark & & & 60.9 & 12.5 & 22.4 & 27.3& 71.4 &86.0 & 89.8 \\
\cmark & & \cmark & 60.9 & 12.9& 22.9 & 27.7 & \textbf{73.8}& 88.2 & \textbf{92.5} \\
\cmark & \cmark & & 60.9 & 13.0& 23.0& 28.2  & 72.9 &87.8 &90.4 \\
\cmark & \cmark & \cmark & \cellcolor{cellgray}60.9 & \cellcolor{cellgray}\textbf{13.7} & \cellcolor{cellgray}\textbf{24.2} & \cellcolor{cellgray}\textbf{29.2} & \cellcolor{cellgray}73.1 & \cellcolor{cellgray}\textbf{88.5} & \cellcolor{cellgray}91.9 \\
\end{tabular}}
\caption{\textbf{Adapters in pre-training.} Pre-training with CLIP Image Adapter (CIA), Image Alignment Adapter (IAA), and Text Alignment Adapter (TAA) is the most effective.}
\label{sec:experiment-ablation:adapters}
\end{subtable}
\hspace{15mm}
\begin{subtable}[t]{0.38\textwidth}
\centering
\resizebox{\textwidth}{!}{%
\begin{tabular}{@{}c|c|ccc|ccc@{}}
\multirow{2}{*}{Stage} & \multirow{2}{*}{\makecell[c]{Img-Txt \\ Acc}} &\multicolumn{3}{c|}{Objaverse-LVIS} & \multicolumn{3}{c}{ModelNet-40} \\
&& Top-1 &Top-3& Top-5 & Top-1 &Top-3& Top-5\\
\midrule
One stage & 55.3 & 12.2 &21.4 &27.0 & 71.4 &86.4& 90.7\\
Two stages & \cellcolor{cellgray}60.9 & \cellcolor{cellgray}\textbf{13.7} & \cellcolor{cellgray}\textbf{24.2} & \cellcolor{cellgray}\textbf{29.2} & \cellcolor{cellgray}\textbf{73.1} & \cellcolor{cellgray}\textbf{88.5} & \cellcolor{cellgray}\textbf{91.9} \\
\end{tabular}}
\caption{\textbf{Pre-training stages.} Learning the CLIP Image Adapter first and then pre-train the 3D encoder with Dual Adapters is better than training modules all together.}
\label{sec:experiment-ablation:stages}
\end{subtable}

\vspace{2mm}

\begin{subtable}[b]{0.32\textwidth}
\centering
\resizebox{\textwidth}{!}{%
\begin{tabular}{@{}cc|ccc|ccc@{}}
\multirow{2}{*}{Image} & \multirow{2}{*}{Text} & \multicolumn{3}{c|}{Objaverse-LVIS} & \multicolumn{3}{c}{ModelNet-40} \\
&& Top-1 &Top-3& Top-5 & Top-1 &Top-3& Top-5\\
\midrule
\cmark & & 11.9 & 20.8 & 25.9 & 67.6 & 87.3 & \textbf{92.5}  \\
& \cmark & 10.5 & 19.2 & 23.7 & 70.2 & 85.0 & 88.0\\
\cmark & \cmark & \cellcolor{cellgray}\textbf{13.7} & \cellcolor{cellgray}\textbf{24.2} & \cellcolor{cellgray}\textbf{29.2} & \cellcolor{cellgray}\textbf{73.1} & \cellcolor{cellgray}\textbf{88.5} & \cellcolor{cellgray}91.9 \\
\end{tabular}}
\caption{\textbf{Pre-training modalities.} Exploiting both image and text data provides the most performance gain. They contribute almost equally in TAMM.}
\label{sec:experiment-ablation:modalities}
\end{subtable}
\hfill
\begin{subtable}[b]{0.32\textwidth}
\centering
\resizebox{\textwidth}{!}{%
\begin{tabular}{@{}cc|ccc|ccc@{}}
\multirow{2}{*}{IAA} & \multirow{2}{*}{TAA} & \multicolumn{3}{c|}{Objaverse-LVIS} & \multicolumn{3}{c}{ModelNet-40} \\
&& Top-1 &Top-3& Top-5 & Top-1 &Top-3& Top-5\\
\midrule
\cmark & & 13.0 & 23.1 & 28.3 & 68.1 & 86.5 & 91.4  \\
& \cmark & 12.9 & 22.5 & 27.5 & 72.4 & 86.2 & 90.4\\
\cmark & \cmark & \cellcolor{cellgray}\textbf{13.7} & \cellcolor{cellgray}\textbf{24.2} & \cellcolor{cellgray}\textbf{29.2} & \cellcolor{cellgray}\textbf{73.1} & \cellcolor{cellgray}\textbf{88.5} & \cellcolor{cellgray}\textbf{91.9} \\
\end{tabular}}
\caption{\textbf{Adapters for inference.} Dual Adapters are complementary, and their combination more accurately classifies 3D shapes at test time.}
\label{sec:experiment-ablation:iaa-taa}
\end{subtable}
\hfill
\begin{subtable}[b]{0.3\textwidth}
\centering
\resizebox{\textwidth}{!}{%
\begin{tabular}{@{}c|ccc|ccc@{}}
\multirow{2}{*}{Images} & \multicolumn{3}{c|}{Objaverse-LVIS} & \multicolumn{3}{c}{ModelNet-40} \\
& Top-1 &Top-3& Top-5 & Top-1 &Top-3& Top-5\\
\midrule
1 & 13.6 & 23.6& 28.9& 72.5& 88.0 & 92.6\\
2 & 13.5 & 23.5 & 28.8 &73.1 &87.6& \textbf{92.8}\\
4 & \cellcolor{cellgray}\textbf{13.7} & \cellcolor{cellgray}\textbf{24.2} & \cellcolor{cellgray}\textbf{29.2} & \cellcolor{cellgray}\textbf{73.1} & \cellcolor{cellgray}\textbf{88.5} & \cellcolor{cellgray}91.9 \\
8 & 13.5 & 23.1 & 28.4& 73.0 & 89.0& 92.8 \\
\end{tabular}}
\caption{\textbf{Number of images.} Aligning a 3D shape with 4 images of it is the best.}
\label{sec:experiment-ablation:image-number}
\end{subtable}
\caption{\textbf{Ablation study of components in TAMM.} We pre-train Point-BERT models on ShapeNet in various settings and evaluate their zero-shot classification performance on Objaverse-LVIS and ModelNet40. The baseline method OpenShape~\cite{liu2023openshape} achieves 10.8\% and 70.3\% Top-1 accuracy on these two benchmarks (Table~\ref{sec:experiment-zsd_1}), respectively. The setting adopted by TAMM is {\sethlcolor{cellgray}\hl{marked}}.}
\label{sec:experiment-ablation}
\vspace{-2mm}
\end{table*}
}

As illustrated in Table~\ref{sec:experiment-few-shot-linear},  
our TAMM achieves the best results and sets a new state of the art in the few-shot classification task. 
TAMM shows \emph{both higher overall accuracy and smaller deviations} than other methods, showcasing the generalizability and robustness of TAMM-learned 3D representations. For example, when pre-trained on the Ensembled dataset, our TAMM surpasses OpenShape~\cite{liu2023openshape} by $2.9\%$, $3.7\%$, $7.7\%$, $5.6\%$, respectively for all four settings. 
These results indicate that our TAMM is able to learn 3D representations that are more generalizable and can be readily adapted to downstream tasks under \emph{low-data regimes}.

\subsection{Real-World Recognition}
To evaluate the capability of TAMM in understanding \emph{3D shapes from the real world}, we follow the previous work CLIP$^2$~\cite{zeng2023clip2} and test TAMM on a real-world recognition task, in which the model aims at correctly classifying each instance from a complex scene in a \emph{zero-shot} manner.
Specifically, we select the real-world scene-level dataset ScanNet~\cite{dai2017scannet} and adopt same data splits as \cite{zeng2023clip2}, containing $17$ classes.
We perform zero-shot classification (Section~\ref{sec:experiment:zero-shot-subsection}) on the point cloud of each object instance extracted from scenes, and report both the Top-1 accuracy of each class and the mean accuracy of all $17$ classes. 
As shown in Table~\ref{sec:experiment-open-world-understanding}, TAMM significantly outperforms all prior methods, showing improvements of $3.8\%$ and $10.9\%$ in average Top-1 accuracy compared with OpenShape and CLIP$^2$, respectively. 
These results underscore TAMM's ability in recognizing and understanding 3D shapes captured from real-world scenarios.
Further results on complex scene recognition and instance segmentation are included in the supplementary material.

\subsection{Ablation Study}
In this section, we provide additional experimental results to further test
the performance gain from each design of TAMM.
Due to limited computation, we \emph{pre-train} Point-BERT~\cite{yu2022point} on the \emph{small-scale} ShapeNet~\cite{chang2015shapenet}, and use two \emph{more challenging} zero-shot classification benchmarks ModelNet40~\cite{wu20153d} and Objaverse-LVIS~\cite{deitke2023objaverse} for \emph{evaluation}. We alter only one component in each set of experiments.

\noindent\textbf{Adapters in Pre-Training.}
As described in Section~\ref{sec:method-clip-tuning}, our CLIP Image Adapter (CIA) mitigates the domain gap between rendered images and natural images.
We first measure the contrastive accuracy, computed as the ratio of image-text pairs where the text feature is more similar to the image from the same triplet than any other text.
The contrastive accuracy of CLIP without CIA is only $40.1\%$, indicating that the original image-text feature space is compromised by the domain gap.
CIA brings the accuracy up to $60.9\%$, showing the effectiveness of our CIA in resolving the shifted data domain. Furthermore, as shown in Table~\ref{sec:experiment-ablation:adapters}, CIA increases the zero-shot accuracy on Objaverse-LVIS dataset from  $12.2\%$ to $13.7\%$, validating its effectiveness on improving the pre-trained 3D representations. 

TAMM adopts Image Alignment Adapter (IAA) and Text Alignment Adapter (TAA) to decouple the feature space and enrich the learned 3D representations.
To further explore the performance gains introduced by our Dual Adapters (IAA and TAA), we experiment tri-modal pre-training with or without IAA and TAA.
As shown in Table~\ref{sec:experiment-ablation:adapters}, both adapters bring performance gains, and their combination achieves the best result, demonstrating the effectiveness of Dual Adapters.

\noindent\textbf{Pre-Training Stages.}
We investigate the significance of our two-stage pre-training design, which initially learns CLIP Image Adapter (CIA) to re-align image-text pairs and subsequently employs Dual Adapters in decoupled tri-modal pre-training. An alternative could be learning these modules all together in one stage.
As shown in Table~\ref{sec:experiment-ablation:stages}, remarkably, the two-stage pre-training approach achieves a better performance compared with one-stage pre-training.

\noindent\textbf{Pre-Training Modalities.} 
We further explore the effectiveness of image and language modalities in pre-training. Table~\ref{sec:experiment-ablation:modalities} shows that integrating the 3D modality with both the image and language modalities consistently yields superior performance across various benchmarks, compared with aligning it with either modality alone.
Moreover,
unlike prior methods which ineffectively learn from images (Figure~\ref{fig:intro-teaser}-left),
the image and language modalities contribute almost equally in TAMM, which also signifies that TAMM better exploits the image modality and brings more gains by learning from both image and text data. 

\begin{figure}[t]
\centering
\vspace{-4mm}
\includegraphics[width= 1 \linewidth]{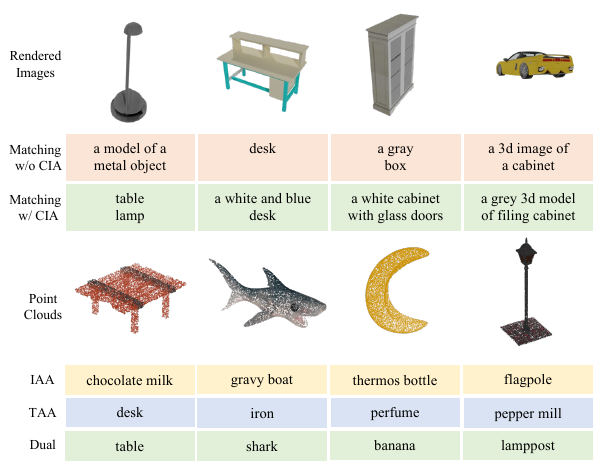}
\caption{\textbf{Qualitative results.} \textbf{Top}: CIA re-aligns the images rendered from 3D shapes with the text descriptions. The rendered images are {\sethlcolor{cellorange}\hl{inaccurately}} matched with text when the image features are directly extracted by CLIP, and CIA can {\sethlcolor{cellgreen}\hl{correct}} the matching. \textbf{Bottom:} IAA and TAA decouple 3D features with complementary visual and semantic focuses. Features from one single adapter are matched with classes whose {\sethlcolor{cellyellow}\hl{appearance}} or {\sethlcolor{cellblue}\hl{semantics}} resemble the true class;
using both adapters leads to the {\sethlcolor{cellgreen}\hl{correct}} class.
}
\label{sec:method-visualization}
\vspace{-2mm}
\end{figure}

\noindent\textbf{Adapters for Inference.}
Our IAA and TAA decouple 3D features with visual and semantic focuses, respectively, offering a more accurate and comprehensive understanding for 3D shapes.
At inference time, we combine the outputs from Dual Adapters for classification tasks.
To investigate this design, we separately utilize features produced solely by either IAA or TAA in zero-shot classification evaluation. As shown in Table~\ref{sec:experiment-ablation:iaa-taa}, the combined features achieve more accurate classification, demonstrating that the learned features from IAA and TAA are complementary to each other.

\noindent\textbf{Number of Images.}
In order to gain more comprehensive knowledge from the image modality, TAMM simultaneously aligns the 3D feature with multi-view 2D images, each projected from a different perspective. We evaluate this design by aligning the 3D feature with a varying number of corresponding 2D images, ranging from 1 to 8 images. As shown in Table~\ref{sec:experiment-ablation:image-number}, multi-view image features offer a performance gain by fully exploiting the image modality. Aligning a 3D shape with 4 views achieves the best result.

\noindent\textbf{Qualitative Results.} Finally, we provide some visualizations to intuitively demonstrate the benefit of our proposed TAMM. As illustrated in Figure~\ref{sec:method-visualization}, CIA successfully mitigates the domain gap introduced by rendered images, and leads to the more accurate matching between images and text. IAA and TAA learn 3D representations with subtly different but complementary focuses on vision and semantics, respectively, and their combination brings more robust and comprehensive 3D representations. More visualized results are included in the supplementary material.

\section{Conclusion}
\label{sec:conclusion}

In this work, we examine the sub-optimal utilization of 2D images in existing multi-modal pre-training methods for 3D shape understanding, and propose TriAdapter Multi-Modal Learning (TAMM), a novel two-stage representation learning approach built on three synergistic adapter modules. Extensive experiments verify that TAMM consistently learns improved 3D features in various settings. 

\noindent\textbf{Limitations.} Due to the limited computation resources, we are not able to perform pre-training on very large-scale 3D backbones with billions of parameters. The quality of the learned 3D representation could be further improved if the 3D backbones are scaled to a larger size. 

\noindent{\footnotesize\textbf{Acknowledgement.} This work was supported in part by NSF Grant 2106825, NIFA Award 2020-67021-32799, the Jump ARCHES endowment, and the IBM-Illinois Discovery Accelerator Institute. This work used NVIDIA GPUs at NCSA Delta through allocations CIS220014, CIS230012, and CIS230013 from the ACCESS program. \par}

\fi

\ifMain
{
    \small
    \bibliographystyle{ieeenat_fullname}
    \bibliography{main}
}
\fi

\ifAppendix
\clearpage
\appendix
\maketitlesupplementary

\ifMain
\else
\addtocounter{figure}{3}
\addtocounter{table}{5}
\addtocounter{equation}{7}
\fi

\section{Implementation Details}
\label{sec:experiment-details}
For a fair comparison with previous methods, we adopt two representative 3D encoders for our study: Point-BERT~\cite{yu2022point} (Transformer-based) and SparseConv~\cite{choy20194d} (convolution-based), following the same architectural configurations as prior methods~\cite{xue2023ulip, liu2023openshape}.
We employ OpenCLIP-ViT-G/14~\cite{cherti2023reproducible} as the pre-trained CLIP model. 
TAMM is pre-trained for $200$ epochs using the AdamW optimizer~\cite{kingma2014adam, loshchilov2019decoupled}, and a cosine learning rate scheduler with and a two-epoch warm-up, and a base learning rate of $5 \times 10^{-4}$.
Regarding the CLIP Image Adapter and Dual Adapters, we set $\alpha$ to $0.2$ in
\ifMain
Equation~\ref{equation-1}
\else
Equation~3
\fi
and employ ReLU~\cite{agarap2018deep} and GELU~\cite{hendrycks2016gaussian} activation functions, respectively, following~\cite{gao2023clip, xue2023ulip}.

\section{Additional Results on Complex Scene Recognition}
To further assess TAMM's capability in understanding 3D shapes from scene data, we conduct experiments using the Hypersim dataset~\cite{roberts:2021}, a photorealistic synthetic
dataset designed for comprehensive indoor scene understanding.  In this experiment, we extract the point clouds of object instances from segmentation annotations and focus on $17$ classes to evaluate TAMM's zero-shot recognition ability. Some classes are excluded due to their amorphous shapes (\eg, ``floor,'' ``ceiling'') or because they are not well-defined for classification (\eg, ``otherfurniture,'' ``otherstructure'').
The results are detailed in Table~\ref{sec:experiment-hypersim}, which demonstrate that TAMM surpasses OpenShape in terms of both overall accuracy and average of per-class accuracy, with respective improvements of $+5.9\%$ and $+1.8\%$. Significantly, TAMM also outperforms OpenShape in $11$ out of the $17$ evaluated classes. This evaluation on Hypersim underscores TAMM’s robustness in recognizing and understanding 3D shapes derived from various scene contexts.

\section{Additional Results on Instance Segmentation}

To delve deeper into TAMM's proficiency in 3D scene understanding, we test whether the 3D backbone pre-trained by TAMM can further enhance SoftGroup++~\cite{vu2022softgroup}, the current state-of-the-art 3D instance segmentation method.
More specifically, we integrate the pre-trained Point-BERT model into the feature extractor module in the top-down refinement stage of SoftGroup++, and subsequently fine-tune the classification branch.
The 3D instance segmentation results on ScanNet are illustrated in Table~\ref{sec:experiment-instance-segmentation}. These results reveal that TAMM can indeed improve the overall performance of SoftGroup++. Notably, TAMM attains an  $AP / AP_{50}$ score of $46.1\%/68.0\%$, marking an enhancement of  $0.6\%/1.0\%$ over SoftGroup++.
Furthermore, as an improved pre-training approach, TAMM exceeds OpenShape by $0.4\%~AP$ and $0.6\%~AP_{50}$.
The results on real-world instance segmentation underscores TAMM's significant potential in the tasks of scene-level 3D understanding.

\begin{table}[!h]
\centering

\resizebox{\columnwidth}{!}{%
\begin{tabular}{c|ccc}
\toprule
Method &  $AP$ &  $AP_{50}$ & $AP_{25}$ \\
\midrule
SoftGroup++\textsuperscript{$\dagger$}~\cite{vu2022softgroup} & 45.5 & 67.0 & 78.7 \\
SoftGroup++ \& OpenShape~\cite{liu2023openshape} & 45.7 & 67.4 & 78.7 \\
SoftGroup++ \& TAMM (Ours) & \textbf{46.1} & \textbf{68.0} & \textbf{79.0} \\
\bottomrule
\end{tabular}}
\begin{spacing}{0.8}
{\footnotesize \textsuperscript{$\dagger$} Reproduced using the original implementation~\cite{vu2022softgroup}.}
\end{spacing}
\caption{\textbf{3D instance segmentation results on ScanNet v2.} Incorportating TAMM into SoftGroup++ improves the $AP$ performance from $45.5\%$ to $46.1\%$, achieving the best results.}
\label{sec:experiment-instance-segmentation}
\end{table}

{
\setlength\tabcolsep{3pt}
\begin{table*}[!t]
\centering
\resizebox{1 \linewidth}{!}{
\begin{tabular}{c|c|c|ccccccccccccccccc}
\toprule
Method &OAvg. & Avg. & Cabi & Bed & Chair & Sofa & Tabl & Door & Bksh & Shlv & Curt & Pill & Clth & TV & Papr & Twl & Nght & Sink & Lamp\\
\midrule
OpenShape\textsuperscript{$\dagger$}~\cite{liu2023openshape} & 56.7 & 48.8 & 13.0 & \textbf{40.0} & 71.2 & 70.7 & \textbf{73.3} & 81.4 & \textbf{20.5} &  54.6 & \textbf{66.7} &  60.0 & 23.9 & \textbf{43.1}  & 36.5 & \textbf{24.0} & \textbf{38.1} & 62.0 & 51.0 \\
TAMM (Ours)\textsuperscript{$\dagger$} & \textbf{62.6} & \textbf{50.6} & \textbf{20.6} & 38.2 & \textbf{75.3} & \textbf{76.2} & 72.6 & \textbf{88.1} & 8.2 &  \textbf{59.1} & 61.9 &  \textbf{62.8} & \textbf{41.0} & 26.2  & \textbf{57.3} & \textbf{24.0} & 31.0 & \textbf{63.0} & \textbf{55.0} \\
\bottomrule
\end{tabular}
}
\begin{spacing}{0.8}
{\footnotesize \textsuperscript{$\dagger$} Results using Point-BERT~\cite{yu2022point} as 3D encoder, pre-trained on the Ensembled dataset.}
\end{spacing}
\caption{\textbf{Zero-shot classification results on the Hypersim dataset.} \textbf{OAvg.}: Overall Top-1 accuracy of all shapes. \textbf{Avg.}: Mean average Top-1 accuracy of all classes. TAMM achieves the best results under both metrics.}
\label{sec:experiment-hypersim}
\end{table*}
}

\begin{figure*}[p]
  \centering
  \includegraphics[width= 0.8 \linewidth]{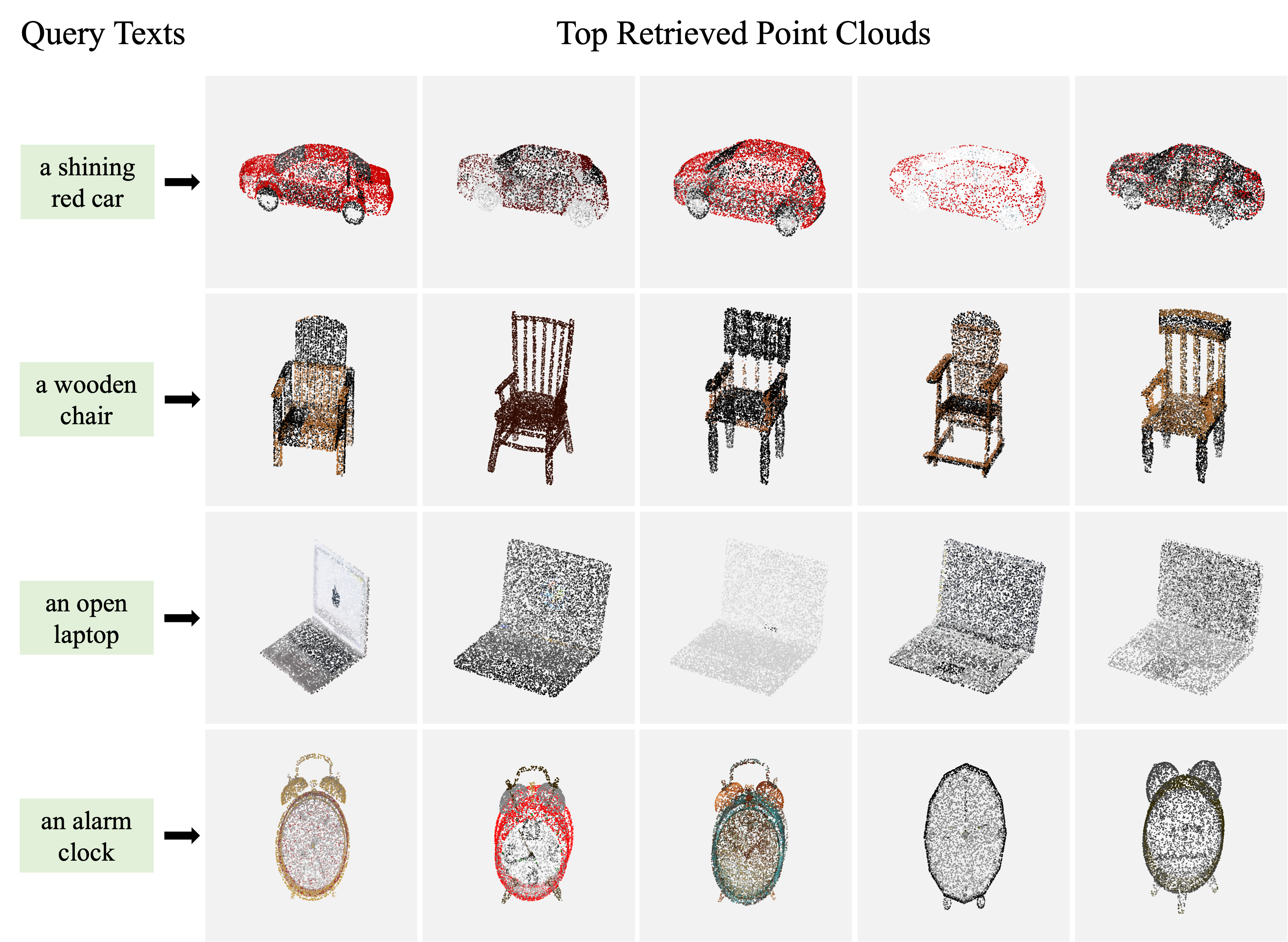}
  \caption{\textbf{Qualitative results of text-to-point-cloud retrieval.} We use TAMM to acquire the features of the given query text and retrieve the point clouds with the most similar features. The shown examples demonstrate TAMM's strong multi-modal comprehension.}
\label{sec:suppl-3D-retrieve-text}
\end{figure*}

\begin{figure*}[!h]
  \centering
  \includegraphics[width= 0.8 \linewidth]{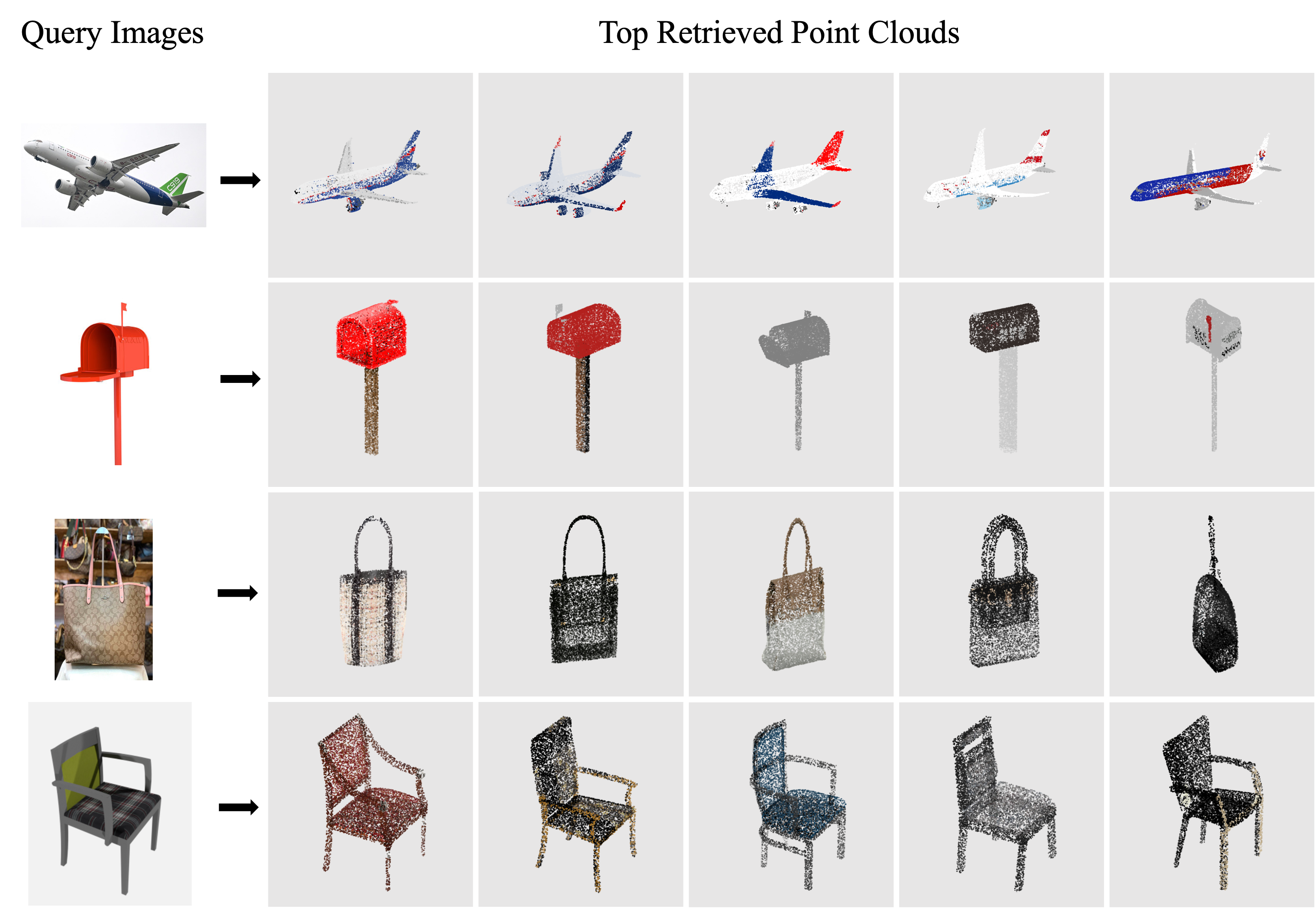}
  \caption{\textbf{Qualitative results of image-to-point-cloud retrieval.} We use TAMM to acquire the features of the given query images and retrieve the point clouds with the most similar features. The shown examples demonstrate TAMM's strong multi-modal comprehension.}
\label{sec:suppl-3D-retrieve}
\end{figure*}

\begin{figure*}[!h]
  \centering
  \includegraphics[width= 1 \linewidth]{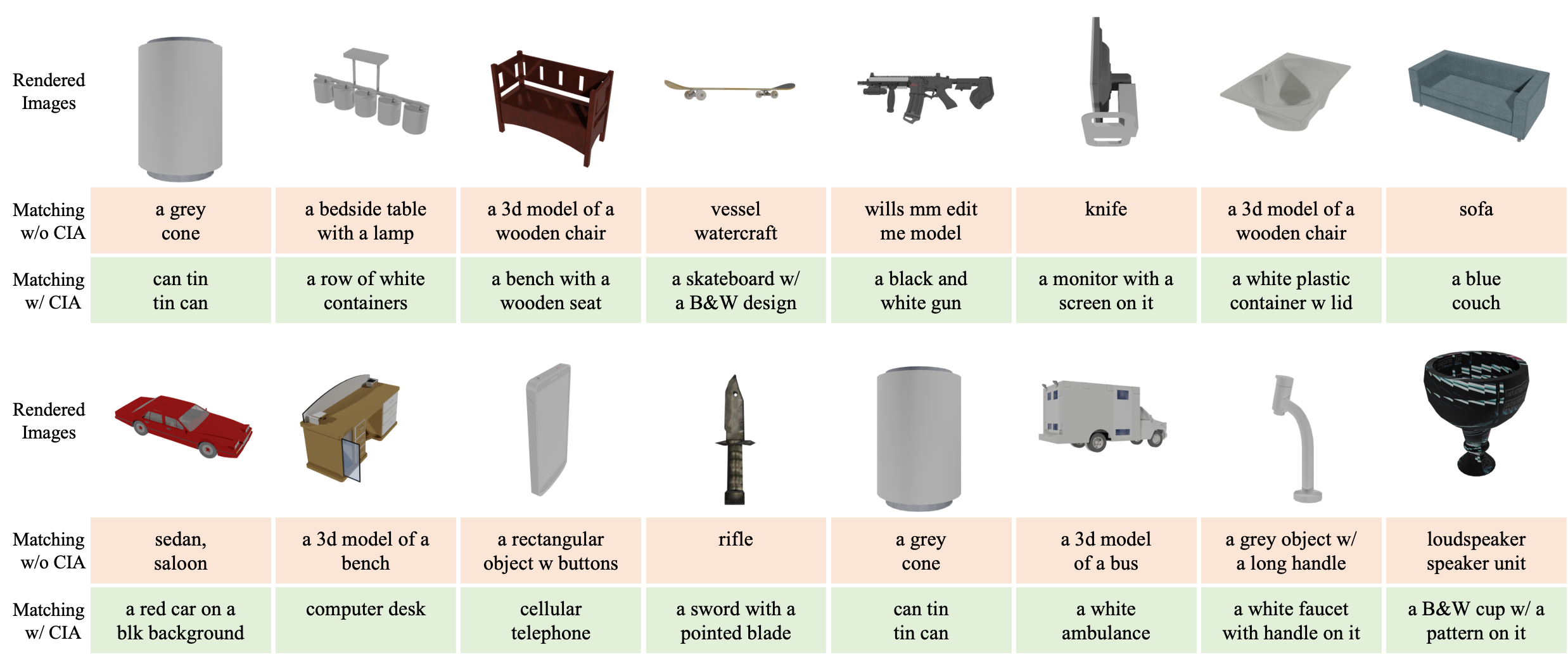}
  
   \caption{\textbf{Qualitative results of CLIP Image Adapter (CIA).} CIA re-aligns the images rendered from 3D shapes with the text descriptions. The rendered images are {\sethlcolor{cellorange}\hl{inaccurately}} matched with text when the image features are directly extracted by CLIP, and CIA can {\sethlcolor{cellgreen}\hl{correct}} the matching. }
   \label{sec:suppl-vis-cia}
\end{figure*}

\begin{figure*}[!h]
  \centering
  \includegraphics[width= 1 \linewidth]{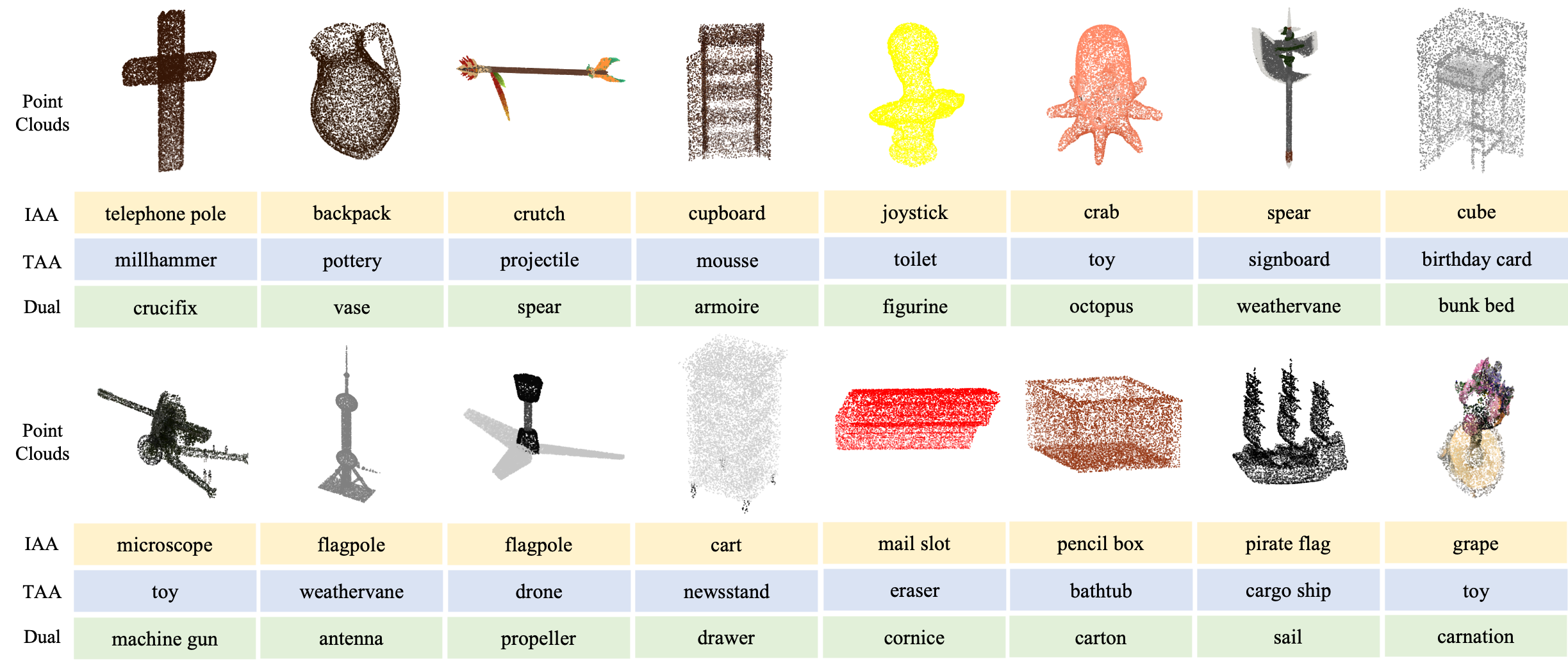}
  \caption{\textbf{Qualitative results of Image Alignment Adapter (IAA) and Text Alignment Adapter (TAA).} IAA and TAA decouple 3D features with complementary visual and semantic focuses. Features from one single adapter are matched with classes whose {\sethlcolor{cellyellow}\hl{appearance}} or {\sethlcolor{cellblue}\hl{semantics}} resemble the true class;
using both adapters leads to the {\sethlcolor{cellgreen}\hl{correct}} class.}
\label{sec:suppl-vis-iaa-taa}
\end{figure*}

\section{Additional Qualitative Results}
\label{sec:additional-qualitative}
In this section, we provide additional qualitative results to supplement the visualizations presented in the main body of the paper. Figure~\ref{sec:suppl-3D-retrieve-text} showcases examples of cross-modal retrieval from text to 3D point clouds.
Figure~\ref{sec:suppl-3D-retrieve} showcases examples of cross-modal retrieval from 2D images to 3D point clouds. More specifically, we extract the adapted features of the query text or image and employ the TAMM-learned 3D backbone to find the point clouds with the most similar features. The retrieved point clouds highly resemble objects in the query text or images, reflecting that the representations learned by TAMM are cross-modal and unified.
Figure~\ref{sec:suppl-vis-cia} demonstrates how our CLIP Image Adapter (CIA) effectively bridges the domain gap caused by rendered images, resulting in more accurate image-text matching. 
Additionally, Figure~\ref{sec:suppl-vis-iaa-taa} illustrates the distinctive yet synergistic roles of Image Alignment Adapter (IAA) and Text Alignment Adapter (TAA). These adapters learn 3D representations with focuses on vision and semantics, respectively. Their integration yields more robust and comprehensive 3D representations, highlighting the effectiveness of our approach.

\fi

\ifMain
\else
\clearpage
{
    \small
    \bibliographystyle{ieeenat_fullname}
    \bibliography{main}
}
\fi

\end{document}